\documentclass[11pt]{article}
\usepackage{authblk}
\usepackage{booktabs} 

 \usepackage[hyphenbreaks]{breakurl}
\usepackage[hyphens]{url}
\usepackage{natbib}
\usepackage[ruled]{algorithm2e} 

\SetAlFnt{\small}
\SetAlCapFnt{\small}
\SetAlCapNameFnt{\small}
\SetAlCapHSkip{0pt}
\IncMargin{-\parindent}

 \newcommand\m[1]{{\color{magenta} Marija: #1}}
 \newcommand\am[1]{{\color{blue} Andreas: #1}}
 \newcommand\ld[1]{{\color{red} Louise: #1}}

 \renewcommand\m[1]{}
 \renewcommand\am[1]{}
 \renewcommand\ld[1]{}

\usepackage{color}
\usepackage{graphicx}
  \usepackage{amsfonts}
    \usepackage{amsmath}

\begin{document}
\title{Towards Moral Autonomous Systems} 
\author[1]{Vicky Charisi }
\affil[1]{%
 University of Twente
}
\author[2]{Louise Dennis }
\affil[2]{%
 University of Liverpool
}
\author[2]{Michael Fisher}
\author[3]{Robert Lieck}
\affil[3]{%
 Universit{\"a}t Stuttgart
 }
\author[4]{Andreas Matthias}
\affil[4]{
 Lingnan University
}
\author[5]{Marija Slavkovik}
 \affil[5]{%
marija.slavkovik@uib.no, 
  University of Bergen
 }

\author[6]{Janina Loh (Sombetzki)}
\affil[6]{%
  University of Vienna
}

\author[7]{Alan F. T. Winfield}
\affil[7]{%
 University of the West of England, Bristol
}

\author[8]{Roman Yampolskiy}
\affil[8]{%
 University of Louisville
}

%
%

%
%


\maketitle

\begin{abstract}
Both the ethics of autonomous systems and the problems of their technical implementation have by now been studied in some detail. Less attention has been given to the areas in which these two separate concerns meet. This paper, written by both philosophers and engineers of autonomous systems, addresses a number of issues in machine ethics that are located at precisely the intersection between ethics and engineering. We first discuss the main challenges which, in our view, machine ethics posses to moral philosophy. We them consider different approaches towards the conceptual design of autonomous systems and their implications on the ethics implementation in such systems. Then we examine problematic areas regarding the specification and verification of ethical behavior in autonomous systems, particularly with a view towards the requirements of future legislation. We discuss transparency and accountability issues that will be crucial for any future wide deployment of autonomous systems in society. Finally we consider the, often overlooked, possibility of intentional misuse of AI systems and the possible dangers arising out of deliberately unethical design, implementation, and use of autonomous robots.
\end{abstract}

 {Keywords: Robot ethics, Machine ethics, Artificial morality, Autonomous systems, Verification, Transparency, Unethical AI}

\thanks{
This article is the result of a series of discussions taken within the scope of the Dagstuhl Seminar 6222 \cite{FisherLSW16}.
Dennis, Fisher and Winfield wish to thank EPSRC for their support, particularly via the ``Verifiable Autonomy'' research project (EP/L024845 and EP/L024861)    
Corresponding Author: Marija Slavkovik,  
University of Bergen, P.O.Box 7802, 5020 Bergen, Norway,  marija.slavkovik@uib.no  
}
 
\section{Introduction}

The so called ``trolley problem" is a  thought experiment introduced in \cite{Foot1967}, whose ethical conundrum continues to fascinate today \cite{wallachBook}. Roughly, it can be summarized as follows: there is a runaway trolley on a railroad track and ahead on the track there are five people tied up, unable to escape being killed by the trolley.  The track splits in two by a lever you control. The lever can divert the trolley on to a second track where there is one person tied up and unable to move. Is it more ethical to divert the train, or let it run its course? 

The emergence of driver-less cars in regular traffic has brought the trolley problem to public attention. Articles such as ``Should Your Car Kill You To Save Others?"\footnote{\url{http://www.popularmechanics.com/cars/a21492/the-self-driving-dilemma/}} are flooding popular science media. It is easy, given the same problem context of traffic, to get sidetracked into confusing solving the trolley problem with controlling the impact driver-less cars will have on traffic and our society as a whole. This, of course,  is not the case. Enabling machines to exhibit ethical behavior is a very complex and very real time-sensitive issue. The driver-less cars are only the forefront of a whole generation of intelligent systems  that can operate autonomously and will operate as part of our society. Which ethical theory to employ for deciding whose death to avoid in a difficult traffic situation is a difficult problem. This, however is not necessarily the most important problem we would  need to solve.  The goal of this position paper is to discuss what does {\em engineering machine ethics} entail.  Once we as a society have discerned what is the right thing for an artificial autonomous system to do, how can we make sure the machine does it? 

 The problem of identifying, discerning, and recommending concepts of right and wrong is the domain of moral philosophy. Moral philosophy, together with the law, act as a system of recommendations regarding which possible actions are to be considered right or wrong (ignoring, for the moment, ethics systems that do not specifically address the morality of individual {\em actions,} e.g. character-based ethics, and which are less useful for the problem at hand).
 
Driver-less vehicles are the most visible of a whole range of technologies. This range includes  also the assisted living technologies, as well as the various embedded decision aid software and solutions. We have been using intelligent systems with  varying degree of autonomy  for the past fifty years: industrial robots, intelligent programming of household appliances, automated trains, etc. However, what all of these machines have in common is that they either operate in a segregated space, a so called {\em working envelope}, or they have no capabilities to cause damage to their environment. Driver-less vehicles are obviously going to ``break" both these restrictions. How can we build intelligent autonomous systems that uphold the ethical values of the society in which they are embedded? This is the main concern of {\em machine ethics}, a new interdisciplinary area of research within Artificial Intelligence (AI) \cite{Moor:2006,1667947,Wallach:2008,wallachBook,AndersonAB16}. 
 
It is clear that choosing the best moral theory to implement in a particular intelligent autonomous system is not a simple question. It is primarily a question for moral philosophy and opens new challenges in this field. In Section~\ref{mp} we give a brief overview of these challenges. 

For a long time, people and societies have been the only intelligent decision-makers. Moral philosophy has been developed with the  often implicit  assumption that the moral agent is a human.  It is not at all clear to what extent existing moral theories extend to non-human decision-makers.  Even if it is shown to be easy to replace a human agent with an artificial agent in a moral theory, and some societal decision is made concerning which ethical behavior in machines is desirable or sufficient, we are still faced with a set of problems regarding the {\em implementation} of ethical reasoning. These are the problems we analyze here: implementation, verification, trust, confidence and transparency and the prevention of intentionally unethical systems. 
 
Moral theories are inherently ambiguous in recommendations of moral behavior, thus reflecting the context dependency of what constitutes a moral choice. We already know that artificial systems, when compared to people, are not good at handling ambiguity. Enabling machines to deal with context-ambiguity in decision-making is a core Artificial Intelligence problem \cite{Russell:2015}. In Section~\ref{approaches} we give an overview of the most intuitive approaches to implementing ethical behavior in autonomous systems and discuss the advantages and shortcomings of these approaches.

Human societies have a multitude of means for ensuring its members behave within the socially accepted boundaries of morality. We can say that a person behaves ethically because they have an individual and personal motivation to do so, without going into how this motivation is formed. The motivation for an artificial agent to behave ethically originates not personally from the agent, but from other actors. These actors can broadly be organized intro three groups: the designers of the artificial agent, its users, and the various societal regulators whose job it is to make sure that order in society is maintained. It is all of these actors that need to be reassured that their own particular motivations for the AI system to behave ethically are met. Hence, a big concern when engineering machine ethics is not only that ethical behavior is accomplished, but also that the ethical behavior can be verified. This is the issue we discuss in Section~\ref{specification-and-verification}.

In Section~\ref{transparency-and-accountability} we focus on issues of transparency and accountability for machine ethics implementations. Since there are several actors outside of the ethical agent who are supplying the motivation for ethical behavior, the implementation of this behavior must be transparent to those actors to the extent and in a manner sufficient for their needs. Transparency is a key element in enabling society to have the right amount of trust and  confidence in the operations of an AI system.  

Lastly in Section~\ref{the-dark-side} we discuss the possible dangers for society that lie in engineering machine ethics. Like all technology, AI systems can also be abused to further criminal activities. AI systems can be deliberately built to behave unethically and illegally, but they also can be induced, deliberately or by accident, into exhibiting socially undesirable behavior. 

The main contribution of this position paper is an integral overview of the immediate challenges and open questions faced when pursuing  the problem of engineering of machine ethics. This paper is a result of a week long discussion among experts of different fields within the scope of the Dagstuhl Seminar 16222\footnote{The report on this seminar is available \cite{FisherLSW16}.}, and incorporates various ideas that arose as a result of discussions among interdisciplinary experts.  Position papers that focus on machine ethics as a whole have been produced and they offer interesting insights in the problem as a whole, see for example \cite{Moor:2006,AndersonA07,Bonnefon2016,AndersonAB16}, but to the best of our knowledge, this is the only document devoted specifically to the problem of engineering machine ethics.

\section{Challenges for Machine Moral Philosophy}\label{mp}
Normative ethics is the branch of moral philosophy concerned with developing methods for deciding how one aught to act. The purpose of devising a moral theory, within normative ethics,  is to develop a consistent and coherent system that can be followed to unambiguously identify an action as good or bad. Numerous specific theories have been developed, the most notable of which are perhaps utilitarianism \cite{Harsanyi1977}, Kantianism \cite{Kant}, and Ross's ethical theory \cite{WDRoss}.  

All of the moral theories so far developed in philosophy have been built around several underlying assumptions.  First and foremost is the   assumption  that the reasoning and decision making when following the moral theory is done by an agent that is a human.  The immediate specific question we can pose to moral philosophy is: given a specific normative moral theory, how is the theory affected if an artificial agent replaces a human agent in it? 

A more subtle assumption in normative ethics is the assumption that the agent  has {\em de facto} the ability of being a moral agent. This means that the theory is developed for a being that is capable of understanding concepts of right and wrong. The debate of whether an artificially created entity can be a moral agent is still far from settled \cite{Etzioni2017}. Wallach and Allen \cite[Chapter 2]{wallachBook} hint at the idea that how {\em ethically sensitive} an artificial system can be depends on how able of autonomous action it is. At this point in the development of the field of machine ethics it is fair to summarise that the capacity for moral agency of an artificial system is more of a sliding scale rather than a Boolean value. There are at least two questions to moral philosophy we can pose. The first is: how to define a scale of moral agency to describe the moral abilities of current and future artificial agents. The second question again is about going back to specific normative moral theories and checking how the theory is affected by replacing a full moral agent with an agent that is ``lower" on the newly devised moral agency scale. Perhaps the concept of moral agency is altogether inadequate when discussing autonomous intelligent systems and a new concept needs to be devised.

There is a further  weaker assumption in normative ethics, which is the assumption  that the agents of the theory are   able to accept and act upon {\em considered judgements} \cite{Elgin1996}. Consider judgments are ``common sense agreements" about what is good or bad in particular instances or cases. For example, the idea that murder is bad except in exceptional circumstances, is a considered judgment. This, let us call it {\em considered judgment ability assumption}, is reflected in the numerous moral dilemmas encountered in the normative ethics literature, the trolly problem being the most popular example. 

A dilemma in  normative ethics, understood in a very broad sense,   is a problem of choosing between two 
options each of which violating one or more considered judgment.  An artificial agent does not possess a common sense understanding unless one is programmed into him. The engineering of common sense, or rather background knowledge, has been a notoriously elusive problem in artificial intelligence. The dilemmas we encounter in machine ethics reflect this. Consider for example the so called Cake or Death problem introduced in \cite{Armstrong15} which describes a situation in which an agent is unsure if  if killing people is ethical or baking them a nice cake  is ethical.
  
Solving Cake or Death is not an ethical problem in moral philosophy, it is trivial. However, the question of which are the essential considered judgments that necessarily have to be implemented in machines is not a trivial problem. This would depend on the nature and abilities of a specific class of artificial agents. In contrast to  focussing on a general theory of discerning right from wrong, there is a need for normative ethics to identify and develop a minimal such theory. 

Building moral machines by implementing human morality is a natural approach. After all, it is human society that those machine are entering and it is human sensibilities and values that they have to uphold. An alternative, or perhaps parallel approach would be to build a normative ethics theory exclusively for artificial agent. An example of such a theory comes to us from science fiction - the three laws of robotics of Prof. Isaak Asimov \cite{Asimov50}. The shortcomings of Asimov's laws of robotics have been extensively argued by the author himself, but they have also been given  a serious philosophical consideration \cite{Anderson08b} and attempts have been made for their implementation \cite{Ethical15,VanderelstW16b,NIK2017}.  

The development of machine moral theory is an interesting open area for study in normative ethics. The question that has to be addressed first perhaps is what constitutes a  viable, desirable, or good theory for artificial agents? We are perhaps primed by the cultural influence of Asimov's laws of robotics to ask: what is it that a robot should never do? But thinking in absolutes is not likely to be viable for machine moral theories any more than it is viable for human normative ethics. Regardless of what comes our of normative ethics in the future, any moral theory developed for artificial agents must be  developed to the point of being {\em implementable}. Prescriptions of good behaviour suffice for people, for machines we need algorithms. An algorithm necessarily includes a specification of  all possible scenarios and context in which  an ethical decision can be made.  Therefore it is necessary that machine normative ethicists,  computer scientists and engineers collaborate closely towards developing machine moral philosophy, with this collaboration perhaps being one of the challenges as well.

\section{Different Approaches, their Advantages and Challenges}\label{approaches}

An intelligent system is one that is capable of communicating with, and reasoning about, its environment and other systems. An autonomous system is one that is capable of, to a certain extent, unsupervised  operation and decision-making.   
Wallach, Allen, and Smit \cite{Wallach:2008,wallachBook} argue that very intuitively,  ethical behavior in machines can be accomplished in at least two different ways. The first approach is to identify a set of ethical rules, perhaps by choosing a normative ethic theory, around which a decision-making algorithm  can be implemented. The second approach is to have a machine evolve or ``learn" to discern right from wrong without it having be explicitly guided by any one ethic theory.   They refer to these two approaches as the {\em top-down} and {\em bottom-up} approach respectively. A hybrid approach in their sense is one in which an agent starts with a set of rules or values and  modifies them into a system for discerning right from wrong.  
  
 Artificial Intelligence (AI) has grown into a large field that incorporates many approaches, which can be, very tentatively, classified into {\em soft computing  approaches}, which include statistical methods, machine learning and probabilistic reasoning, and {\em traditional symbolic AI methods}, which includes logic-based reasoning \cite{Russell:2015}. The question of how to  implement machine ethics in an intelligent autonomous system necessarily hinges on the AI methods that system uses. Different AI approaches would be subject to different machine ethics implementations and we need to consider their malleability to machine ethic approaches, as well as their risks and advantages in this respect.  
 
We here roughly classify the current and future machine ethics implementations based on the main AI approach used into {\em soft machine ethics} and {\em symbolic machine ethics} mirroring the two largest traditional branches of AI methodology.  We discuss both of these approaches and their advantages and challenges. We should note at this point that a hybrid approach here would be one that combines symbolic methods, such as for example rule based reasoning, with soft methods such as for example  supervised learning. 

The  bottom-up approach of \cite{Wallach:2008,wallachBook}  naturally lends itself to be approached by using soft computing AI methods, whereas their top-down approach is perhaps best ``served" by symbolic AI methods. 


\subsection{Soft and Symbolic AI Methods for machine ethics}

Within engineering, a top-down approach towards solving a task consists in breaking down the task iteratively into smaller sub-tasks until one obtains tasks that can be directly implemented. Problems best solvable by a top-down approach are ones in which the problem, and its context, are fully understood and can be formally specified.  This is normally the case when the problem occurs in a controlled environment.  These problem properties are also ones required for a successful solution by implementation using symbolic AI methods such as rule based reasoning.  

There are numerous ways in which a symbolic AI approach can be taken to develop ethical behaviour in a system. The most frequent in the literature is to constrain the choices of the system using rules derived from an ethical theory. This is the approach taken in  \cite{ArkinUW12},  for developing the concept of {\em ethical governor}, and  also in \cite{Ethics:RAS:2015} where the ethical theory used is a version of  Ross's ethical theory \cite{WDRoss}. The \cite{Ethics:RAS:2015} work considers a hybrid autonomous system three-layer architecture: a continuous system controlled by a rational software agent, which makes discrete decisions, via a continuous control layer  that allows for a dynamic environment interaction and feedback. The rational software agent is provided with an ethical policy, a total order over abstract ethical principles such as ``do no harm", ``do not damage property" etc.  The agent relies on external entities to identify if, and which, of her possible actions impinges on some of these abstract ethical principles. Having her actions annotated, the agent chooses between possible actions by selecting the one that is minimally unethical with respect to the given ethical policy.  In contrast, \cite{BendelAnnotatedMachines} proposes a method for building {\em annotated decision trees} for making simple ethical choices.


A bottom-up approach to problem solving  in engineering starts with  describing instances of desired solutions by using adequate parameters and the proceeding to build up a procedure for identifying solutions based on these parameters. Machine learning methods in AI take a bottom-up approach to problem solving.    There are several examples  of using machine learning to implement machine ethics, such as for example \cite{AndersonA14} and\cite{abel_reinforcement_2016}.
In \cite{AndersonA14}  inductive logic programming is used over a corpus of particular cases of ethical dilemma  to  discover ethical preference principles. Each case  relates two actions, one more ethical than the other. The preference between the actions depends on  ethically relevant features that actions involve such as harm, benefit, respect for autonomy, etc. Each feature is represented as an integer that specifies the degree of its presence (positive value) or absence (negative value) in a given action.  The system is able to extract an ethical rule from the cases it is presented with and thus to a certain extent is able to learn to discern right from wrong.  In \cite{abel_reinforcement_2016} reinforcement learning is used to learn what the most moral of two actions is, by rewarding the ``correct" decisions an agent makes and ``punishing" the bad ``wrong" ones.

Whether a soft or symbolic AI method is used depends on the nature of the problem that needs to be solved.  The two families of methods tackle  problem  from different sides and are not mutually exclusive. Each of the methods comes with its own  advantages and challenges with respect to building ethical behaviour in an intelligent autonomous system.  

%

\subsection{Advantages and challenges of using Symbolic AI Methods}

Symbolic AI methods are  best suited for ethical reasoning in limited domains when the context of the decision-making problems can be predicted.  To use symbolic based reasoning,  an ethical theory needs to chosen before constructing and deploying the system and this theory does not change throughout the system's usage. This allows for a thorough and well-informed process of decision-making and the verification of the system prior its practical application. Different ethical principles and theories may explicitly be encoded into the system giving clear options to decide upon. Any parameters that are left open for definition by the customer or to be learned from interaction with the environment have a clear function and it is possible to verify that they do not violate more general ethical principles.


Symbolic AI methods  also come  with their set of challenges and limitations. General ethical guidelines are typically formulated on a very abstract level. Much philosophical discourse on ethics is concerned with problems occurring when applying such general guidelines to concrete situations. It is thus not clear if, and how, general ethical guidelines can be
leveraged to solve concrete problems of decision making in practice. For a real-world system the connection to the non-discrete sensory-motor level must be made. There are many ways to transform continuous sensor values into discrete propositions that can be used in reasoning. General guidelines or even single terms and concepts are only (if at all) implementable in a reduced way, i.e. restricted to one preferably clear interpretation. Due to their context sensitive definition it is not possible to consider every possible interpretation of an abstract guideline or term in implementing them in an artificial system \cite{matthias2011}.

Furthermore, symbolic AI approaches risk conflicts between the implemented ethical theories and principles. 
If only one theory is implemented, e.g. Kant's Categorical Imperative \cite{Kant} or Isaac Asimov's first Law of Robotics\footnote{The laws can be found quoted in Wikipedia, at \url{http://en.wikipedia.org/wiki/Three_Laws_of_Robotics}}, then  this theory would  determine  the specific maxims that are to be defined situationally by the artificial system. Winfield \cite{Winfield2014} describes experimental trials of a minimally ethical robot which implements Asimov's three laws of robotics. The chosen theory must be such as to allow implementable rules to be  derived from it. Such a monistic approach assumes that there exist no moral dilemmas, i.e. that the implemented theory is able to give a conflict-free rule to make a decision in every context (\cite{Winfield2014} experimentally shows how a single ethical rule performs when faced with a balanced ethical dilemma). 

Deciding on a specific set of ethical principles involves settling long-standing
philosophical disputes in an ad-hoc way. It is possible that governmental bodies might take the lead in outlining high-level ethical guidance to designers. For example, Germany's ministry of transport recently announced the intention to set out a basic ethical policy to be followed by car designers stipulating that  property damage takes always precedence over personal injury, that there must be no classification of people, for example, on size, age and the like, and that -- ultimately -- it is the manufacturer who is liable\footnote{\url{http://www.wiwo.de/politik/europa/selbstfahrende-autos-dobrindt-gruendet-ethikkommission-fuer-automatisiertes-fahren/14513384.html}}.   


Symbolic AI approaches require that any kind of   ``common sense" is explicitly coded using the formal language of the method. This severely impacts the scalability of  the machine ethics solution. As it is well understood in the AI sub-discipline of knowledge representation and reasoning, the more expressive the formal language for encoding knowledge is, the more computationally expensive reasoning becomes. It is not a problem of having or symbolically encoding a large amount of information, but a problem of computing   logical entailment or consistency, which are in the core of deep reasoning and are known to be  of non-deterministic computational complexity.   
Therefore, it is unsurprising that the most recent  major AI breakthroughs  have been achieved using statistical processing of information and shallow reasoning. That being said, some symbolic-methods, such as model checking, are scalable within reasonable parameters, and have been vastly deployed in the information processing industry.

\subsection{Advantages and challenges of using Soft AI Methods}

Soft AI methods are best applicable when we know the kind of data the AI system receives
from interacting with the environment and, while the overall objective might not
be well known or specified, we still have an idea how to process these data in a
useful manner. For instance, the vision pipeline of a household robot is
designed to extract obstacles (walls, tables, etc.), objects of interest (books
on a shelf etc.), and its own position from the sensory data because this
information is useful for a wide range of tasks it will be required to perform.
Most real-world AI systems will be partly designed using a soft AI method  at
least on the lower sensory-motor level. Because soft AI methods are based on synthesis of actions and choices, with respect to the task of building an
ethical AI  the major question here is:  how do components designed in a bottom-up fashion affect the overall
ethical properties of the system?

Soft AI methods do not require predetermines moral principles , ethical theories or sets of rules, but instead formulate basal parameters and intend to implement competences whereby an artificial system acts autonomously. This can be done, for example,  via trial and error or other modes of learning such as imitation, induction and deduction, exploration, learning through reward, association and conditioning \cite{Cangelosi:2014}.
Soft AI methods can be separated into models of evolution \cite{Froese:2010} and models of human socialization \cite{Fong2003,Breazeal:2002}. The former simulate evolutionary moral learning, by assessing slightly different programs in an artificial system to evaluate an ethical case. Those programs that can solve the ethical task sufficiently go through to a ``next round" where they are (re)combined to solve further ethical tasks. Evolutionary approaches can be used in earlier stages of moral development before considering models of human socialization.
 
Models of human socialization consider the role of empathy and emotion for moral learning.  They assume that a robot learns morality via empathy \cite{Slote2007}.  What is controversial in the philosophical discourse is that there exist two types of empathy \cite{Stueber2006}: perceptual empathy, when an emotion triggers an equivalent or congruent reaction in the observer \cite{Misselhorn2009}, and imaginative empathy that requires a change in perspective in the form of empathising with the other, putting oneself in the observed other's shoes. Perceptual empathy is explicable with the help of specific {\em theories of mind} or neuronal resonance and mirror neurons and has been implemented in a rudimentary fashion in artificial systems \cite{Balconi2012,Rizzolatti2008,Mataric:2000}.  Ekman \cite{Ekman1992} implements perceptual empathy in the form of a basal affect program as an autonomous reaction scheme as a route to the implementation of morality in robots.   Young children and chimpanzees are equipped with this fundamental form of empathy which forms the basis for pre-social behavior \cite{Warneken2009,Hoffman2000}. 
 Imaginative empathy is much more complex and develops on the basis of perceptual empathy only.  It is exhibited only in human socialisation, not in non-human primates. This form of empathy is cognitively more ambitious and is involved in more complex moral reasoning and acting \cite{Gallagher2012}.  We are not aware of any attempt to implement imaginative empathy in artificial systems.

Since, by means of a soft AI solution the  AI system becomes a moral agent (if only in the narrowest sense of the word) one might ask whether it is likely to be more adaptable to making ethical choices in situations that are not pre-determined (which is a strong limitation to using the symbolic AI methods). Since the system learns its own ethical rules, it circumvents, to an extent (one could argue), the need to choose one particular ethical theory to implement. But this seems at least questionable. Every self-learning system must still be configured to pay attention to particular {\em features} of the data set, and to ignore others. Looking at the {\em consequences} of an action, instead of the agent's {\em motivation} (for example) is such a choice of features that essentially determines the choice of moral theory. It seems difficult to judge at this point whether we can hope to create ethical-theory-agnostic AI systems, since every choice of relevant data features is already, to some extent, a choice of moral theory.

 A major challenge with using soft AI methods is that it is hard to certify whether
the system fulfils any requirements one might want to impose.  Indeed this is a challenge for all machine learning systems.  A machine learning solution virtually behaves as a black box - the approach solves a problem successfully  most of the time, but it is unclear whether a solution can be expected for sure, or why a particular solution was learned or developed for a particular problem.  Soft AI methods, and machine learning in particular, have had a dramatic success recently, with machine learning methods being used in a variety of problems and contexts. This success has prompted for calls to ensure that some level of explainability for the choices of the system is required, which in turn have given rise for the Explainable AI  (XAI) DARPA programme \footnote{\url{https://www.darpa.mil/program/explainable-artificial-intelligence}}. In \cite{AndersonA15} we find one of the earliest specific implementations of XAI in machine ethics. Their system  extracts a moral rule from a collection of cases and is able to explain why a particular decision is identified as more ethical referring to the learning data.

Nonetheless, the black box nature of soft AI methods is likely to mean that these solutions  are unsuitable for implementation in critical systems. This fundamental
problem occurs irrespective of whether the ethical system itself or only
low-level sub-systems are built using a soft AI solution.

\subsection{Modular and Hybrid approaches}\label{hybrid-approaches}
Both the soft and symbolic AI methods   come with advantages and challenges, but they also can complement each-other.   A  system is an entity comprised of several entities, thus in principle an AI system can be built  using components that exploit both solution approaches. We are unaware of any implemented hybrid ethical reasoning   system\footnote{ Although the title of \cite{vanRysewyk2015} claims a hybrid implemented system upon closer inspection it is not clear in which sense is the solution not a pure bottom-up approach in the sense of \cite{wallachBook}, while the authors themselves do not offer an analysis  of this type in their paper.}, but we can very briefly discuss some recommendations for how such a system can be created.

One approach would be to separate decision-making by the ethical principles it involves. For example, decisions involving the possibility of human death should be made using a pre-programmed ethical policy, while decisions involving violation of autonomy can be based on ethical preferences learned through interaction with the system's owner. Another approach would be to separate decision-making in different contexts, with soft AI methods   being allowed as the default ethical decision-making method, while symbolic AI  approaches being required to be   implemented for certain specific pre-determined contexts.   
Alternatively a system can be designed so it first learns to recognize the ethical implications of its actions and then those implications can be used to follow an implemented ethical theory when choosing some specific course of action. 

Implementing ethical reasoning within a system is not sufficient, we must execute such implementation in a way that allows for verification of the quality of the resulting ethical behavior. The designers and manufacturers of AI systems necessarily have to  offer reasons for their users to trust the artificial ethical system,  and they also need to foresee possible malfunctions and provide means to deal with them.

\section{Specification and Verification of  Ethical Behavior}
\label{specification-and-verification}

Within our society,  entities that  are in a position to do us harm, be it a complex machine production tool,  the surgeon operating on our unconscious body, the other drivers on the highway, or a chainsaw, are subject to licensing and certification.  Certification informs consumers and experts of the properties of a product, a system, or a person in a position of responsibility. Knowing that a standard has been met allows individuals to have confidence in using machinery and to trust the decisions and actions of professionals. Tools and systems are certified to operate within designated parameters, while under (well defined) proper care. Certification confirms that the manufacturer has taken all steps necessary to avoid or minimize foreseeable risks that arise in relation to the usage of the tool.  Certification for persons in position of responsibility is more complex because it involves a (possibly continuous) examination to demonstrate that the certified person has the understanding and skills necessary to perform his/her duties. Typically, this involves regulations prescribing \emph{expected} behavior --- often, humans must pass an examination concerning these regulations. Once we move to an autonomous system, with no human directly in control, what are our means to ensure that a systems actually matches the relevant criteria?  

In order to be  confident in a system's behavior we need to \emph{specify} what we can expect the system to do in a particular circumstance, \emph{verify} that the system does actually achieve this, and \emph{validate} that our requirements are actually what the end-users want. There exist a vast range of different techniques, for example developed over many years within the field of \emph{Software Engineering}~\cite{Sommerville01:book}. These techniques range from the \emph{formal}, such as proof, through \emph{structured}, such as testing, to \emph{informal}, such as user validation. All these approaches can, in principle, be applied across the range of autonomous systems, including robotics~\cite{FDW:CACM13}.

\subsection{Who is the confirmation of ethical behavior for?}
\label{Who-is-the-confirmation-of-ethical-behavior-for}

What constitutes an appropriate specification and verification methodology for ethical behavior depends on who is to use the results. In the case of intelligent autonomous systems at least three interested parties can be discerned: the designers including developers and engineers working on developing and maintaining the systems, the end-users, owners or customers, and lastly various government and trade regulatory bodies and insurance agents. Although these three categories are the evident interested parties, this issue of interest discernment is an open problem in its own right, and as some preliminary investigations show\footnote{\url{http://robohub.org/should-a-carebot-bring-an-alcoholic-a-drink-poll-says-it-depends-on-who-owns-the-robot/}} finer discernment may be required.

Although those actually constructing the AI system  may have an intimate knowledge of its internal workings, it is still important that developers and engineers not only have confidence in their prototypes but have techniques for highlighting where issues still remain. The technology itself should not be a black box, but should be open to maintenance and analysis, and must be flexible enough to be improved dynamically.

For end-users, customers and owners, the primary concern is that the AI system they interact with is safe and behaves ethically with respect to the ethical norms they themselves follow, as long as these are within the scope of what is considered ethical and legal within their society.  \emph{Trust} is a key issue and, in order to have trust extended to AI systems, the user needs to be informed of its range of capabilities. The future of AI systems and their proper integration within our society is subject, paradoxically, to undue levels of both optimism and pessimism in terms of the extent to which people can trust such systems. Close attention must be paid to nurturing the appropriate level of trust. 

AI systems are an exciting technological development that have long been anticipated as part of the future in various works of fiction and there is the temptation to play-up their apparent capabilities, particularly by early  marketing when the producers are still seeking financiers for their products. This could lead to the customers placing an unwarranted level of trust in some technology, even when adequate disclaimers and use guidelines are outlined by the manufacturer, which can in turn lead to disastrous consequences\footnote{\url{https://www.theguardian.com/technology/2016/jun/30/tesla-autopilot-death-self-driving-car-elon-musk}}. Such misplacement of trust is dangerous for users in the present, and may cause society to over-react in order to limit integration of technologies which given proper time to adequately develop  would have been advantageous to the same society.  

The appearance of trustworthiness is similarly an issue when people interact with an AI system. For example, a robot might \emph{appear}  ``experienced,'' ``benevolent,''  or ``sympathetic''.  Such appearances are of particular concern for  AI systems that are integrated in assisted living technologies.   Concerns have been raised with respect to the impact assisted living technologies can have on the  elderly \cite{Sharkey2012}. Similarly, it has been shown that children who interact with robots derive expectations of them and ascribe abilities to them.   We need to develop an understanding of the potential  long-term effects of robots on child development \cite{matthias2015}. \m{Need reference to studies involving children and robots.}

Trust should play a considerable role in choosing an ethical theory to implement in AI systems. The ethical theory that is easiest to implement may not necessarily be the one that is most trusted by society.  This was demonstrated in the case of utilitarianism and driver-less cars \cite{Bonnefon2016}.

It is important to note that \emph{trust}~\cite{ArkinUW12} is not equal to ethics. Trust is a social construct intimately concerned with how each individual views the behavior of a robot or system. There may well be some varieties of \emph{objective} trustworthiness, but there will remain many varieties of \emph{subjective} trustworthiness. Many items affect users' level of trust~\cite{SalemLAD15}, for example, the relationship between \emph{trust} and \emph{harm}.  If you could show that robot causes no harm, would you trust it more?

Those who must regulate AI systems and their integration within society also need \emph{confidence} in the system. In addition, the Insurance industry needs to be clear where responsibility~\cite{Janina:responsibility} lies and so where \emph{liability} lies. The concept of liability is likely to be complex and may be split over several actors, such as the manufacturers/designers, the operators and the environment. \emph{Regulation} is crucial and first steps have been taken to go beyond safety and reliability regulations~\cite{BrysonW17,standard:iso13482} into considering the ethical aspects that should be taken into account~\cite{standard:bsi8611}.
 
Any system operating in the real world would eventually find itself in a situation in which it will malfunction and AI systems are no exception. The question is thus how certain can one be in the verified ethical behavior of an AI system and what measures can be taken to mitigate  the consequences of, and learn from, a system's potential failure. This is the issue of having confidence in the system.
%

 In terms of safety standards, the ``gold standard'' is currently that of aircraft autopilot software where safety is measured as the number of accidents per miles flown. We might think that, for AI systems, at least as much confidence is needed. But, is the standard too high or even achievable? There are of course, noticeable differences between aircraft and other autonomous systems. While the operational environment of an aircraft is very controlled and limited, as the aircraft must adhere to a strictly defined flying corridor, the severity of accidents is very high. {\em E.g.,} any malfunction in the air is certainly fatal for all of the aircraft passengers measuring in the hundreds, whereas a miscalculation on the road does not need to be fatal since cars carry fewer passengers than airplanes. It is possible that the hours of operation per accident alone is not always the best measure to assess safety of an autonomous system, but that the severity of the damage caused and the number of individuals involved in an accident should also be taken into account~\cite{UAS:Cert:2015,KellyM01,DenneyP14,JACIC13}.
 
 \ld{This whole section seems rather unfocused.  It starts by talking about who has an interest in the system, then spends several paragraphs talking about trust, briefly considers confidence, liability and regulation (all the space of a few sentences), before talking about accidents per mile and a comparison with the avionics industry.  Is it possible to split it somehow? or provide a clearer thread of argument.}



\subsection{\emph{What} do we want the system to do?}
\label{what-are-the-expectations-what-do-we-want-the-system-to-do}
A key problem is specifying \emph{what} our expectations of an AI system are. Although this is beginning to be codified where safety is considered, for example through \emph{robot safety} standards\footnote{See~\cite{ISO:TC299} for a range or robotic safety standards.}, it is less clear where the ethical/moral requirements should come from and in what form should they be represented? The BS8611 standard~\cite{standard:bsi8611}, for example, does not prescribe \emph{what} the ethical requirements should be, but maps out the issues over which ethical decisions should be considered.  

An obvious route for ethical and legal requirements is  through regulatory or standards bodies. These  entities have the ability to set overall standards, potentially with the help of domain experts. In addition, designers may well have built in specific  ethical codes that go beyond (though do not contradict) those prescribed by regulations. Finally, the user herself may wish to input her ethical preferences, ensuring that the AI   acts in a way that is personally acceptable. Since there are multiple actors that need to define and refine the ethical requirements of the system, each with varying levels of technical expertise, the issue arises of how the ethical requirements are represented for the machine and for concerned actors.  No one clear methodology  emerges. One possibility is to have them represented   in the form of a set of legal or formal rules, as argued in \cite{SaptawijayaP16}. Another possibility is to use  a set of example scenarios developed to test specific ethical choices, as in \cite{AndersonA14}. A third, but by no means final, possibility is as a statistical envelope around a large (and possibly random) set of test cases, against which the AI system must be exhaustively assessed.


\subsection{\emph{How} do we show that the AI systems meets  the
expectations?}
\label{how-do-we-show-that-the-robot-fits-the-expectations}
There is a well-established body of work tackling the \emph{Verification and Validation} (V\&V) of systems, both hardware-centred and software-rich. The aim of \textit{Verification} is to ensure that a system meets its
requirements; \emph{Formal Verification} takes this further, not only having precise formal requirements, but carrying out a comprehensive mathematical analysis of the system to `prove' whether it corresponds to these formal requirements. There are many varieties of formal verification, the most popular being \emph{model checking}~\cite{Clarke00:MC,ArmstrongGLOPRW12}, whereby formal requirements are checked (usually automatically) against \emph{all} possible executions of the system. Verification, via model checking, is widely used especially for the analysis of the \emph{safety} and \emph{reliability} of robotic systems both in terms of physical navigation~\cite{MitschGP13} and in terms of internal decision-making~\cite{EASS:vern}.  What is being verified is that the behaviour of a particular system conforms to defined expectations. In terms of ethical/moral verification, it seems clear that if an AI system acts by following mathematically specified rules, we can potentially formally verify its high-level behavior. Only recently, however, has the use of formal verification for \emph{ethical} or \emph{moral} issues begun to be addressed~\cite{Ethics:RAS:2015,Ethical15}. 

A practical alternative to fully formal verification is to use sophisticated \emph{coverage-driven analysis} methods, appealing to Monte-Carlo techniques and dynamic test refinement in order to systematically ``cover'' a wide range of practical situations. Especially where real-world interactions and devices are involved, testing is likely to be crucial. Indeed, testing for safety and reliability of robotic systems is well-established~\cite{MossigeGM15}. Such model-based testing is a well-developed technology but, as
we move to more complex (ethical) issues sophisticated extensions may well be required. Though such approaches are typically used \emph{before} deployment, related techniques provide a basis for run-time verification and compliance testing~\cite{RosuH05}.  Testing is not as exhaustive as formal proof, but can cover many more scenarios. 

\emph{Validation} is the process of confirming that the final system has the intended behavior once it is active in its target environment, and is often concerned with satisfying external stakeholders. For example, does our system match ethical standards or legal rules set by regulators? Does our system perform acceptably from a customer point of view, and how well do users feel that it works~\cite{lehmann2013should}?  There are
many approaches to carrying out validation, typically involving the assessment of accuracy, repeatability, trust, usability, resilience, etc. All must be extended to cope with ethical and moral concerns.

It is clear that the strength and breadth of V\&V research should allow us to extend and develop this towards ethical and moral concerns. However, a number of issues remain, as follows.
\begin{itemize}
\itemsep1pt\parskip0pt\parsep0pt
\item If the core software is not purely rule-based, for example involving some sub-symbolic learning procedures, then we will need a symbolic representation of the learned content if we are to carry out formal verification of the above form. One of the limitations of both formal verification and testing is likely to be in verifying learning procedures, especially where new ethical principles and preferences of behavior are learned. 
\item Fully formal verification is likely to be unrealistic for complete, complex systems both because of non-symbolic  components (as mentioned above) and because of practical complexity limits.
\end{itemize}
However, we can formally verify \emph{parts} of the system under particular circumstances. There are things that can be proved about core parts of \emph{the system} and about the system's \emph{outputs}. Consequently, formal verification techniques can provide \emph{some} evidence. In assessing how much confidence we need in the V\&V of AI system ethics, it may be possible to leave the burden of this decision to the regulator, manufacturer or end-user as appropriate. So long as a clear indication of the extent of the V\&V of a system exists a user or other interested may take the decision about the risk involved in using the system. Note that we can potentially separate regulation from verification and so allow a variety of different V\&V techniques to be applied.

Lastly, we would like to include here the existing efforts of validating a system that uses soft AI methods, which is  the discussion of {\em Ethical Turing Tests}. Ethical Turing Tests were introduced in \cite{AllenVarnerZinser2000}.  In \cite{AndersonA14} this idea is further fleshed out and implemented. Under an ethical Turing test, both the AI system and an ethicist resolve the same dilemmas.  The system passes the test if its choices are sufficiently similar to the ones of the ethicist. Whether a variant of a Turing test is a sufficient indicator of a certain type of ``human-like" behaviour from a machine is a topic that has been argued as long as the artificial intelligence field exists. All the issues that have been raised, and exhaustively discussed in artificial intelligence, against the Turing test original can be argued to hold for an Ethical Turing test.

\section{Transparency and
accountability}\label{transparency-and-accountability}

The opacity and transparency of deep neural network algorithms has become a major research subject area in recent years. Without knowing how the algorithm functions concerns about inherent biases in data and algorithm itself create difficulties in either discerning forensically or through explainability how decisions were made. Furthermore, there are questions as to whether the actual logic used by learning systems can be explained to people, or whether any explanation would be created after the fact.

The problem with interpretability/explainability is that in some cases it may be impossible to provide a complete and accurate explanation for how a black box system has arrived at its decision. As complexity of systems increases, it is not unusual for the algorithm to extract a million-dimensional feature vector and assign unique weights to each feature. Any human-readable explanation for the decision will include some top N most important features and completely ignore the rest. A human-comprehensible explanation canât be too long or too complex. A good metaphor for this is how we explain things to children if they are not old enough to fully appreciate nuances of the problem. âWhere do kids come from?â âYou buy them at the store!â Any human-friendly explanation from a sufficiently complex system has to be a partially inaccurate simplification or a complete lie.    

The choice of the relevant criteria for an AI system to be deemed ethical will eventually need to be taken by society as a whole. Therefore \textit{transparency} is of utmost importance and thus ensuring transparency is a major challenge. To this end it is necessary to identify \emph{what} has to be transparent to \emph{whom}, and \emph{how} this can be realized. 

Transparency is a key requirement for ethical machines. Important attributes flow from transparency including \textit{trust}, because it is hard to trust a machine unless you have some understanding of what it is doing and why, and \textit{accountability}, because without transparency it becomes very difficult to understand who is responsible when a machine does not behave as we expect it to\footnote{Although it is important to note that transparency is not the same as accountability.}. An ethical machine will need to be transparent to different stakeholders in different ways -- each suited to that particular stakeholder. In this section we consider the transparency needs of a range of stakeholders before considering aspects of transparency common to all. This section   outlines how and why transparency is important to four different groups of stakeholders: users, regulators (including accident investigators), ethicists/lawyers and society at large. Each group has different transparency needs, some of which will have to be met by allowing an AI system's ethics, and ethical logic, to be human readable, or through public engagement. Other needs will require new human-robot interfaces.

Some literature exists on the topic of transparency in AI and autonomous systems. Owotoki  and Mayer-Lindenberg  \cite{Owot2007} proposes a theoretical framework for providing transparency in computational intelligence (CI) in order to expose the underlying reasoning process of an agent embodying CI models. In a recent book Taylor and Kelsey \cite{Tayl2016} make the case for the importance of transparency in AI systems to an open society. For autonomous robots \cite{Wort2016} describes early results showing that building transparency into robot action-selection can help users build a more accurate understanding of the robot. There is also no doubt that transparency is high on policy agenda: the 2016 UK Parliamentary Select Committee on Science and Technology's final report on Robotics and AI expresses concerns over both decision making transparency and accountability and liability\footnote{\url{http://www.publications.parliament.uk/pa/cm201617/cmselect/cmsctech/145/145.pdf}}. Indeed the EU's new General Data Protection Regulation, due to take effect as law in 2018, creates a ``right to explanation'' such that a user will be able to ask for an explanation of an algorithmic decision that was made about them \cite{Good2016}.

\subsection{Transparency to the user}\label{transparency-to-the-user}

Although the critical importance of the human-machine interface is well understood, what is not yet clear is the extent to which an ethical machine's ethics should be transparent to its user. It would seem to be unwise to rely on a user to discover a machine's ethics by trial and error, but at the same time a machine that requires its user to undergo a laborious process of familiarisation may well be unworkable. 

For care robots for instance it may be appropriate for the user to configure the ``ethics" settings (perhaps expressing the user's preference for more or less privacy) or, at the very least, allowing the user to choose between a small number of ``preset" ethics options. There is of course always some danger that many users will rely on the default setting. What is clear is that how these options are presented to the user is very important \cite{matthias2015}; they should for instance help and guide the user in thinking about their `value hierarchy'. The robot might for instance explain to the user ``what would happen" in different situations and hence guide their preferences \cite{Theo2016}. 

For other robot types, driverless cars for instance, the ethics settings may be fixed (perhaps by law) and therefore not user configurable. However, the need for the user to understand how the car would behave in certain situations remains critically important -- especially if the car's design (or the law) requires her to act as a safety driver and assume manual control when the autopilot cannot cope. Even for fully autonomous cars in which the user is only ever a passenger, the person with legal responsibility for the car should be aware of the car's ethics settings. For fully autonomous cars there should still be some user interface so that the passenger can discover, or perhaps ask for help, if the vehicle become unexpectedly immobile or starts behaving erratically.

\subsection{Transparency to regulatory bodies}\label{transparency-to-regulator}

It is clear that the ethics of ethical robots  needs to be transparent to those responsible for (i) certifying the safety of ethical machines, and (ii) accident investigators. Both regulators and accident investigators will be working within a governance framework which includes standards and protocols. The role of the protocols is to set out how robots are certified against those standards, and -- following an accident -- how the accident is investigated.

\emph{Regulators} will need the ability to determine that a machine's ethics comply with the appropriate standards\footnote{Standards Which do not yet exist.}, and making such a determination will require those ethics to be coded and embedded into the robot in a readable way. We might imagine something like a standard Ethics Markup Language (EML -- perhaps based on XML) which codes the ethics. The EML script would be embedded in the robot in a way that is accessible to the regulator, noting that the script will need to be secured to prevent attack from hackers.

\emph{Accident Investigators.} When serious accidents happen, as they inevitably will (see Section \ref{Who-is-the-confirmation-of-ethical-behavior-for}), they will need to be investigated. To allow such investigation, data must to be recorded, suggesting the need for a robot equivalent of the flight data recorder. Therefore an \textbf{Ethical Black Box} (EBB) is proposed  \cite{WinfieldJ17}-- a device that records all relevant data including, crucially, internal state data on the robot's ethical governor. Although the data stored by the EBB would be vital for investigating all aspects of an accident, including causes unrelated to the robot's ethics, here we are interested in accidents which might have been caused by a fault or deficiency in the robot's ethics programming. By recording the sequence of internal states of the ethical reasoning in the moments before the accident, the EBB would allow an investigator to discover exactly why the robot made an incorrect decision. Information that would be important both in determining accountability, and to make recommendations for upgrading the robot's ethics and prevent the same accident happening again.

Specifying the EBB is beyond the scope of this paper, this work has been carried further in \cite{WinfieldJ17}. It is however clear that research is needed to determine what data the EBB must record, the frequency and time window of that data, and how the privacy of that data is maintained. One thing we can be sure of however is the need for an industry standard EBB (as in the aviation industry).  Different EBBs and EBB standards will of course be needed for different applications, but for driverless cars for instance, a single standard EBB should be mandated. Such an EBB would itself require an industry standard, and protocols for certification, fitting and maintenance of EBBs.

\subsection{Transparency to ethicists /
lawyers}\label{transparency-to-ethicists-lawyers}

A third group of stakeholders includes lawyers, who might be required to advocate for AI systems' owners, or on behalf of anyone who makes a claim against an  AI system's owner, or ethicists who might, for instance, be required to act as expert witnesses, in a court of law. If we consider an accident in which a robot's ethics are implicated (see Section \ref{transparency-to-regulator} above), it is clear that both lawyers and ethicists will need to understand (i) a robot's ethics, (ii) the process the robot uses to make an ethical decision (in other words how its ethical reasoning works), and (iii) the data captured by the ethical black box. Providing this kind of transparency to lawyers and ethicists will not only be necessary, but is also likely to be challenging, as robot manufacturers and designers may regard such details, especially (ii), as proprietary IP.

Another category of expert stakeholder includes psychologists, who might be required to either evaluate robots for their potential to cause psychological harm to the user, or as expert witnesses, in providing an investigation with an expert evaluation of the psychological harm caused to robot user(s) in a particular case.

\subsection{Transparency to the whole of society}
\label{transparency-to-the-whole-of-society}

AI systems -- and especially ethical AI systems -- are a disruptive technology, with potentially significant societal and economic impact, thus an easily overlooked but important stakeholder is society as a whole. We only need to consider driverless cars and trucks to appreciate the level of potential disruption, to jobs and transport policy for instance, as already reflected in the level of public and press interest in this technology.

It is therefore very important that the ethics of ethical AI systems should be transparent to society at large, for two reasons. First, because citizens should be able to make informed judgments about the kinds of AI system they wish to have in their lives, and even more importantly those they do not want in their lives, so that they can lobby their elected representatives and ensure that government policy properly reflects those views. And second, if society is to have confidence in the ethics of a class of ethical AI system (driverless cars, for example) then it should accept a degree of collective responsibility for those ethics.




%
%
%

\subsection{Technical means to bring about transparency\label{technical-means}}

It is clear that the different stakeholders outlined above have very different transparency needs. Some of those needs are met through making the ethical rules and logic readable (for instance for regulators, ethicists or lawyers), but for others transparency can only be met through technical means. Here we briefly outline several approaches to meeting those needs.

\begin{itemize}
\itemsep1pt\parskip0pt\parsep0pt
\item
Assisted living AI systems would benefit from a ``Why did you do that?'' button which, when pressed, causes the robot to explain -- perhaps using speech synthesized text -- why it carried out the previous action. We could call the system behind this an ``explanation module''. For an AI system with a fixed set of responses the explanation module should be relatively easy to implement, but for an AI system which learns its ethics such an implementation could be challenging; in either case the explanation module and its user interface would need very careful design in order to meet the needs of a non-technical user. 
\item
  An ethical AI system which makes use of simulation based internal models as part of some ethical governor (for example \cite{Winfield2014}) might allow us to go further than the ``Why did you do that?" button, by making the robot's internal simulation accessible to the user. This would enable the user to ask the robot ``What would you do?'' in a given situation. Clearly such a facility would need a much more sophisticated user interface than a button press, but through visualisation tools we can imagine the user watching the robot's internal simulation running through various scenarios on a connected laptop or tablet device. Note that a similar visualisation interface would be of great value to accident investigators (Section \ref{transparency-to-regulator}), and expert witnesses or lawyers (section \ref{transparency-to-ethicists-lawyers}) to \textit{play back} a robot's internal simulation in the moments leading up to an accident, and what the alternatives open to the robot at the time might have been.
\item
 The technical requirements for an ethical back box (EBB) were already outlined in Section \ref{transparency-to-regulator} above.
\end{itemize}

\section{Dangerous and Deliberately Unethical AI}\label{the-dark-side}
  Finally, it is important to be aware of the ways people may abuse or manipulate AI systems. 
 As with all technology AI systems can also be deliberately abused for malice or to further one's  illegal goals \cite{yampolskiy2016taxonomy}. While our primary concern is to contribute towards designing  AI systems that behave ethically within a human society \cite{sotala2015responses,yampolskiy2015artificial} and promote human and animal welfare, some concern also needs to be raised about how that AI system can protect itself against abuse \cite{yampolskiy2016artificial}. By abuse we, of course, do not mean mistreating the AI system in the sense in which a person or an animal can be mistreated, but taking advantage of the capabilities and opportunities offered by the AI system to commit criminal acts. 
 
 The abuse of an AI system can be achieved by hacking an existing system or by deliberately creating an unethical AI system \cite{pistono2016unethical,Vand2016}. Hacking itself can be accomplished in several ways. The code of the AI system might be directly hacked.  But a system can also be manipulated by interaction and such manipulation does not necessarily require technical knowledge.  This is illustrated by the short-lived Tay experiment. Tay was an artificial intelligence chatter-bot released by Microsoft Corporation on March 23, 2016 and taken offline 16 hours after launch\footnote{\url{http://www.bbc.com/news/technology-35890188}}. Tay was programmed to learn from conversation, however it took the netizens a very short time to ``train''  it into making morally questionable statements. 
 
 Manipulation by interaction can be accomplished both deliberately and by accident.  A learning based system can be led \cite{yampolskiy2014utility} intro eliciting bad conclusions through crafted case descriptions, etc. By this means one can slowly train systems away from moral behavior. As an example of accidental manipulation consider the example of children learning that driverless cars slow down in their presence, they might choose to make a game out of it. Children playing with car's  reactions might annoy passengers by causing delay; and might ultimately lead to the disabling of safeguards.\ld{Is this actually an example of manipulation?  It's not the system that is being manipulated here. ROMAN - I say yes, the system is being "indirectly" manipulated}
Purposeful creation of Malevolent AI can be attempted by a number of diverse agents with varying degrees of competence and success. Each such agent would bring its own goals/resources into the equation, but what is important to understand here is just how prevalent such attempts will be and how numerous such agents can be.  For example, we should be concerned about: the \emph{military} developing cyber-weapons and robot soldiers to achieve dominance; \emph{governments} attempting to use AI to establish hegemony, control people, or take down other governments; \emph{corporations} trying to achieve monopoly, destroying the competition through illegal means; \emph{villains} trying to take over the world and using AI as a dominance tool; \emph{black hats} attempting to steal information, resources or destroy cyber infrastructure targets; \emph{doomsday cults} attempting to bring the end of the world by any means; the \emph{depressed} looking to commit suicide by AI; \emph{psychopaths} trying to add their name to history books in any way possible; \emph{criminals} attempting to develop proxy systems to avoid risk and responsibility; AI \emph{risk deniers} attempting to demonstrate that AI is not a risk factor and so ignoring caution; and even AI \emph{safety researchers}, if unethical, attempting to justify funding and secure jobs by purposefully developing problematic AI.

%
%
%
 The ethical and unethical behaviors of an AI system are not necessarily symmetrical.   Existing systems define only a small part of the problem space \cite{yampolskiy2015space}.  Apart from ethical and unethical behavior, an AI system can also exhibit a  behavior that has neither been programmed nor predicted as a particular combination of otherwise ethical rules and choices. 

Lastly we must mention the potential for ``cultural imperialism" when designing the ethical behavior of an AI system. With globalisation, a product's production and consumers are diverse. What constitutes ethical behavior in one region may even be considered unethical in another. All the involved actors, the designers, users and society, both on the supplier and on the demand end of the AI system need to be aware of the reality that the supplier society ethics influences  the ethical behavior of the AI system, which in turn   influences the ethics of the society in which the AI system operates.   

\section{Summary}
Moral philosophy has a very rich history of studying how to discern right from wrong in a systematic, consistent and coherent way. Today we have a real need for a \textit{functional} system of ethical reasoning as  AI systems that function as part of our society are ready to be deployed. Building an AI system that behaves ethically is a multifaceted challenge. The questions of which ethical theory should be used to govern the AI system's behavior has received most of the attention. Here, we focus on the problem that comes next, after what is right or wrong for a machine to do is decided -- how to implement the ethical behavior. 

The problem of engineering ethical behavior is made complex because of the prime motivators for such behavior. For humans, the motivation for behaving ethically is primarily internal. Without falling into difficult philosophical arguments on the existence of free will, we accept that people are capable of behaving ethically because they choose to do so, although, of course, they too can be motivated towards ethical behaviour by incentives, punishments and assignment of liability. For AI systems, however,  the motivation towards ethical behaviour is exclusively external because it can always be traced back to their design and it cannot be reinforced in the same way as it can be done with people. This motivation furhtermore comes from several stakeholders. We cannot claim that the full list of these stakeholders can even be known before the AI systems are fully deployed, but we can discern between the three most evident groups of stakeholders: the designers, the users and the various regulatory organs of society. Each of these stakeholders needs to play their role in deciding what is the best ethical behavior for a given AI system, but they also need to be convinced in an adequate way that the implemented behavior actually yields the desired results. A moral AI system needs to be adequately transparent and accountable to each group of stakeholders.   

Unlike people, who more or less share the same ``hardware" and reasoning capabilities, machines and AI systems can be built using many different approaches.  The implementation of ethical reasoning will depend not only on what the stakeholders need and desire, but also on what is possible given the chosen problem-solving implementation. We discussed the two basic implementation approaches reflecting two large families of AI methods:  the soft AI and the symbolic AI families, and identify the challenges and advantages of each. 

An AI system capable of ethical behavior is necessarily a complex system. With complex systems two things are evident: that they will malfunction and that they can be used to attain criminal goals. We discuss methods of verifying that an AI system behaves as designed within specified parameters, but we also discuss how the engineering of the ethical behavior impacts available options once a system malfunctions. Lastly we discuss in broad strokes what the stakeholders need to be aware of in terms of abuse of an AI system with ethical behavior capabilities, both when that abuse is intentional and accidental.

\bibliographystyle{named}
\bibliography{group234}

\end{document}